\documentclass[letterpaper]{article} 
\usepackage{aaai25}  
\usepackage{times}  
\usepackage{helvet}  
\usepackage{courier}  
\usepackage[hyphens]{url}  
\usepackage{graphicx} 
\usepackage{natbib}  
\usepackage{caption} 
\usepackage{subcaption}
\usepackage{multirow}
\usepackage{multicol}
\usepackage{color}
\usepackage[noend]{algpseudocode}
\usepackage{algorithm}
\usepackage[dvipsnames]{xcolor}
\usepackage{colortbl}
\definecolor{mygray}{gray}{.93}
\definecolor{mygray2}{gray}{.87}
\usepackage{amssymb}
\usepackage{amsfonts}
\usepackage{booktabs}
\usepackage{pifont}
\newcommand{\cmark}{\ding{51}}%
\newcommand{\xmark}{\ding{55}}%
\usepackage{makecell}
\usepackage{dsfont}
\usepackage{enumitem}
\usepackage{newfloat}
\usepackage{listings}
\DeclareCaptionStyle{ruled}{labelfont=normalfont,labelsep=colon,strut=off} 
\floatstyle{ruled}
\newfloat{listing}{tb}{lst}{}
\floatname{listing}{Listing}
\title{Prior-Constrained Association Learning for \\ Fine-Grained Generalized Category Discovery}

\author{
    Menglin Wang\textsuperscript{\rm 1},
    Zhun Zhong\textsuperscript{\rm 2}\thanks{Corresponding authors.},
    Xiaojin Gong\textsuperscript{\rm 3}\footnotemark[1]
}
\affiliations {
    \textsuperscript{\rm 1}School of Computer and Electronic Information, Nanjing Normal University, China\\
    \textsuperscript{\rm 2}School of Computer Science and Information Engineering, Hefei University of Technology, China\\
    \textsuperscript{\rm 3}College of Information Science and Electronic Engineering, Zhejiang University, China\\
    lynnwang6875@gmail.com, zhunzhong007@gmail.com, gongxj@zju.edu.cn
}

\begin{document}

\maketitle

\begin{abstract}
This paper addresses generalized category discovery (GCD), the task of clustering unlabeled data from potentially known or unknown categories with the help of labeled instances from each known category. Compared to traditional semi-supervised learning, GCD is more challenging because unlabeled data could be from novel categories not appearing in labeled data. Current state-of-the-art methods typically learn a parametric classifier assisted by self-distillation. While being effective, these methods do not make use of cross-instance similarity to discover class-specific semantics which are essential for representation learning and category discovery. In this paper, we revisit the association-based paradigm and propose a Prior-constrained Association Learning method to capture and learn the semantic relations within data. In particular, the labeled data from known categories provides a unique prior for the association of unlabeled data. Unlike previous methods that only adopts the prior as a pre or post-clustering refinement, we fully incorporate the prior into the association process, and let it constrain the association towards a reliable grouping outcome. The estimated semantic groups are utilized through non-parametric prototypical contrast to enhance the representation learning. A further combination of both parametric and non-parametric classification complements each other and leads to a model that outperforms existing methods by a significant margin. On multiple GCD benchmarks, we perform extensive experiments and validate the effectiveness of our proposed method. 
\end{abstract}

\begin{links}
\link{Code}{https://github.com/Terminator8758/PAL-GCD}
\end{links}

\section{Introduction}
\label{sec:intro}
The success of deep learning models has mostly been driven by the availability of large-scale annotated datasets. However, it is costly and inefficient to annotate all the data, especially as datasets grow larger in an open world. Semi-supervised learning~\cite{oliver2018realistic} thus emerges to be a promising direction for learning with both labeled and unlabeled data. Typical semi-supervised learning assumes unlabeled data comes from known categories. Nevertheless, data from novel categories frequently appear in the real world, limiting the applicability of semi-supervised learning. As a relaxation to this assumption, generalized category discovery (GCD) has been proposed~\cite{vaze2022generalized}, allowing unlabeled data to belong to both known and unknown categories. The target of GCD is to recognize images from both old and new categories by learning a model that clusters unlabeled images into distinct semantic groups, making it more practical for discovering novel categories with the assistance of old category data. 

Current methods explore the GCD task from two perspectives, representation learning and parametric classification. The initial GCD paper~\cite{vaze2022generalized} uses self-supervised contrastive learning to learn robust representation, removing the need for parametric classifier. The problem is that it overlooks the intrinsic semantic relations among samples, causing the learned representation to be less discriminative. Indeed, samples belonging to the same potential category call for attraction instead of general repulsion. Later methods~\cite{pu2023dynamic, zhao2023learning, zhang2023promptcal} exploit cross-instance similarity relations to discover semantic groups or \textit{k}-NN positives, and such grouping result guides the contrastive learning towards finding a discriminative feature space. However, the quality of grouping is determined by the design of data association strategy, and severe noise can be incorporated if the association design is not reliable enough. As an alternative, SimGCD~\cite{wen2023parametric} revives parametric classifier through self-distillation learning and entropy regularization, which has become a popular baseline. But as the parametric classifier is implicitly regularized, its weights may not capture class-specific semantics well. Also, self-distillation alone may not provide strong enough supervision for the distinction of different classes.

\begin{figure*}[ht]
\centering
\includegraphics[width=0.95\textwidth]{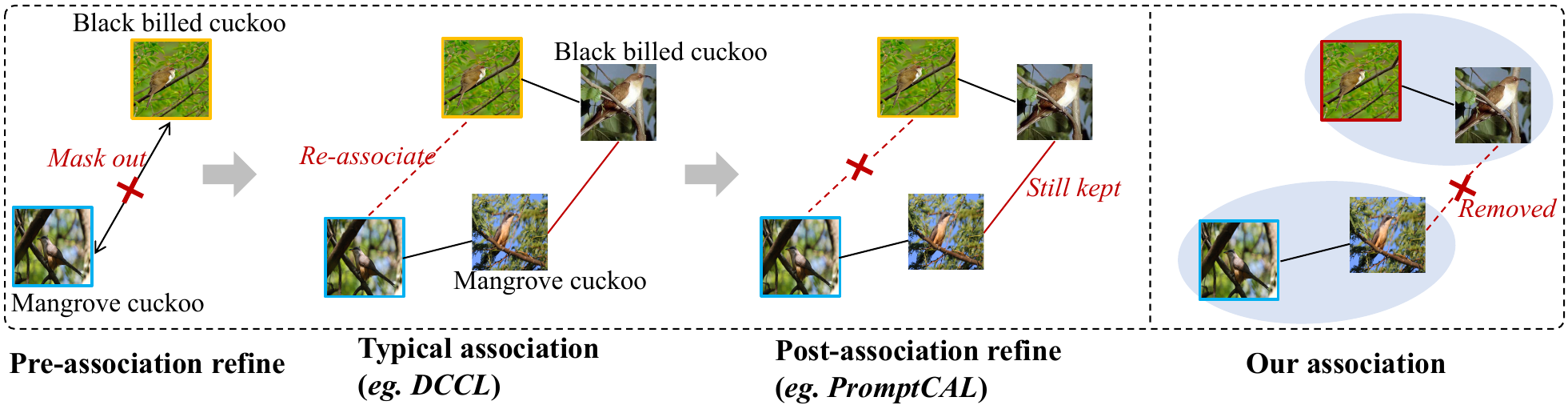}
\caption{An illustration of implicit false association, and how our proposed association avoids it. Arrowed line represents masking out the distance, solid line indicates direct association of two instances, and dashed line denotes indirect association. Red line indicates false association. Images with colored edges indicate they are from labeled subset.}
\label{fig_intro}
\end{figure*}

In this paper, we re-examine the previous association-based GCD methods, and identify their weakness in the association design that makes their performance inferior, especially on fine-grained datasets. The aim is that through better association design, more accurate instance groupings can be estimated to facilitate model representation learning. Specifically, Figure \ref{fig_intro} provides an intuitive example to illustrate the limitations of previous association designs and our motivation. As shown in the figure, some methods utilize the labeled prior for pre-association refinement, i.e. masking out the distance of images from different known categories. However during the association, the masked image pair could be re-connected by indirect association of other images. Such association will generate inaccurate groupings of labeled and unlabeled instances. Even if a post-association refinement removes the connection of labeled images, the unlabeled images' association is still kept, causing undesirable groupings to appear.

In light of this, we aim to circumvent such possibly emerging situations by fully incorporating the labeled prior into the association process. As Figure \ref{fig_intro} shows, our association keeps track of the updated image groups. When a new image pair appears, we not only examine the image pair, but also check their corresponding groups to determine if the image pair should be associated. In the example, the two groups contain known yet label-conflicting instances, therefore the image pair will not be associated. In this way, both direct and indirect false association can be avoided. By repeating the instance-wise association steps,  more faithful instance groupings can be obtained.

With the groupings generated by association, we adopt non-parametric prototypical contrastive learning to strengthen the representation. The association based non-parametric classification serves as a stand-alone and good-performing framework. Specially, the prior knowledge of ground truth class number is not required, which makes the framework flexible in practical applications. Additionally, we also propose to optionally perform association on a subset of all samples to improve the scaling of the association to more data scenarios. We explore the possibility of combining parametric classifier with the proposed association based parametric classifier, which is proven effective when the number of classes is known. Through a two-stage training pipeline, we exploit their complementarity and unify them in one framework to mutually boost each other. This unified model achieves strong performance on multiple datasets.

Our main contributions are summarized as follows:
\begin{itemize}[noitemsep,topsep=0pt]
  \item We propose a simple yet effective method for fine-grained GCD. By re-examining the role of association, a novel prior-constrained association algorithm tailored for GCD task is proposed. 
  \item With the assistance of proposed association, we unify non-parametric and parametric classification under one single framework, where representation learning and classifier learning can mutually boost each other.
  \item Extensive experiments on both fine-grained and generic datasets demonstrate the effectiveness of the proposed method. Compared to previous best method, our method improve the accuracy by $4.4\%$ and $15.3\%$ on CUB and Stanford Cars respectively.
\end{itemize}

\section{Related Work}
\textbf{Generalized Category Discovery (GCD)} draws similarity with novel category discovery (NCD) in both containing labeled and unlabeled images and aiming to discover novel categories in the unlabeled set. Initial GCD method~\cite{vaze2022generalized} learns representations by self-supervised contrastive learning on all data, along with supervised contrastive learning on labeled subset. SimGCD~\cite{wen2023parametric} constructs an effective baseline using parametric classifier. A later variant $\mu$GCD~\cite{vaze2023norep} improves SimGCD by using a teacher network to provide supervision for self-augmented image pairs. More recently, SPTNet~\cite{wang2023sptnet} learns spatial prompts as an alternative to adapt data for better alignment with the model. DCCL~\cite{pu2023dynamic} proposes to mine sample relations by generating dynamic conceptions using improved Infomap clustering~\cite{rosvall2008infomap}, followed by conception and instance-level contrastive learning. Similarly, GPC~\cite{zhao2023learning} also estimates prototypes by Gaussian mixture model and a split-and-merge to take labeled instances into account. PromptCAL~\cite{zhang2023promptcal} improves the ViT backbone by learning auxiliary prompts, as well as affinity propagation on KNN graph to estimate instance relation. Although labeled data is exploited to assist clustering in these methods, it is often taken as a pre- or post-clustering refinement. As such, the potential benefit of labeled instances are not fully exploited.  As a comparison, we fully incorporate the labeled data prior during every step of the association process, empowering reliable association of unlabeled data by taking advantage of the labeled instances as bridges.

\noindent\textbf{Prototypical Contrastive Learning (PCL).}
In recent years, contrastive learning~\cite{Gutmann2010NCE} has proven as an effective technique for self-supervised learning~\cite{wu2018memory, he2020momentum, Chen2020SimCLR, Li2021pcl} and other settings~\cite{Khosla2020SCL, Wang2021ICS, zhao2023learning}. In particular, prototypical contrastive learning compares instances with a set of prototypes encoding class-specific semantic structure, leading to discriminative embedding space. As such, many vision tasks have exploited PCL for method design. ProtoNCE~\cite{Li2021pcl} combines instance-wise contrastive learning and multi-grained PCL for transfer learning. \cite{ge2020self,Wang2021CAP} adopt iterative clustering based PCL for object re-ID. A few methods~\cite{pu2023dynamic, zhao2023learning} in GCD have also considered PCL to learn discriminative representation. The critical issue for prototypical contrast is how to obtain representative prototypes, which then comes down to designing effective association strategy. Our method also adopts PCL, however, our better utilization of prior and design of semi-supervised association lead to more reliable prototypes, which in turn facilitates learning better representation.

\noindent\textbf{Data Clustering and Association.}
Clustering has long been used as a way to discover potential semantic groups within the data. Unsupervised clustering methods like K-Means~\cite{hartigan1979algorithm}, DBSCAN~\cite{ester1996density} and hierarchical clustering~\cite{johnson1967hierarchical, murtagh2012algorithms} are widely used in many applications~\cite{ge2020self, wang2022hierarchical, pu2023dynamic}. Semi-supervised clustering is also studied in some works~\cite{bair2013semi, bilenko2004integrating}. Basu \textit{et al.}~\cite{basu2002semi} propose constrained K-Means by enforcing that labeled instances are assigned to their own cluster during K-Means iteration. COP-Kmeans~\cite{wagstaff2001} modifies K-Means to make sure no constraints are violated when assigning instances. Constrained DBSCAN~\cite{ruiz2010density} and hierarchical clustering~\cite{davidson2005clustering} are also considered. Metric-based methods~\cite{yin2010semi, klein2002instance, xing2002distance, lange2005learning, pu2023dynamic, zhang2023promptcal} modify the pairwise distance such that two instances with a "must-link" constraint have a lower distance, and those with a "cannot-link" constraint have a larger distance. Our proposed association is also constraint-based, but the constraints are enforced during a threshold-based group merging process, during which new categories are allowed to be discovered.

\section{Methodology}
\subsection{Overview}
Under Generalized Category Discovery setting, we consider the problem of clustering images in a dataset among which a subset has known class labels. Assume the dataset $\mathcal{D}$ is comprised of two parts $\mathcal{D_L}=\{(x_i, y_i)\}_{i=1}^N \in \mathcal{X}\times \mathcal{Y_L}$ and $\mathcal{D_U}=\{(x_i, y_i\}_{i=1}^M \in \mathcal{X}\times \mathcal{Y_U}$, where $\mathcal{D_L}$ is the labeled subset of $N$ images whose labels $\mathcal{Y_L}$ are known, and $\mathcal{D_U}$ is the unlabeled subset of $M$ images whose labels $\mathcal{Y_U}$ are not known. Image labels in $\mathcal{D_U}$ is a superset of image labels in $\mathcal{D_L}$, i.e. $\mathcal{Y_L}\subset \mathcal{Y_U}$. Given dataset $\mathcal{D}$, the aim is to correctly recognize and cluster the images in $\mathcal{D_U}$ containing known and unknown categories. To address the GCD task, we seek to improve the representation learning by estimating reliable semantic groups as the guidance. To this end, we incorporate the labeled data prior into the association process, and design a prior-constrained greedy association algorithm. Such association generates faithful instance groups as well as class-representative proxies to guide the model representation learning. 
Finally, to exploit the synergy of non-parametric (prototypical contrastive learning) and parametric classification, we unify them in one framework by joint two-stage optimization.

\subsection{Limitation of Previous  Methods}
\label{sec:3.2}

There have been some recent attempts~\cite{pu2023dynamic, zhang2023promptcal, kim2023proxy, zhao2023learning} at estimating the semantic structure by semi-supervised clustering or association, so as to provide stronger and explicit supervision to unlabeled data. Albeit with acceptable performance, we take a closer look at the current association designs in GCD and discover that there exists missing clues and the association can be further optimized with the given labeled data prior.

In GCD task, it is natural to utilize the labeled data from known categories to assist the association of unlabeled instances. Current association-based methods usually adopt the labeled data as a pre- or post-clustering refinement. For pre-clustering refinement, after computing the pairwise distance of instances, those between known yet different categories can be directly masked as disconnected. Then the refined distance matrix would be input to a standard clustering algorithm~\cite{rosvall2008infomap, ester1996density}. For post-clustering refinement, after unsupervised clustering, the associations between known different categories are removed as a refinement.

\begin{figure*}[ht]
\centering
\includegraphics[width=0.85\textwidth]{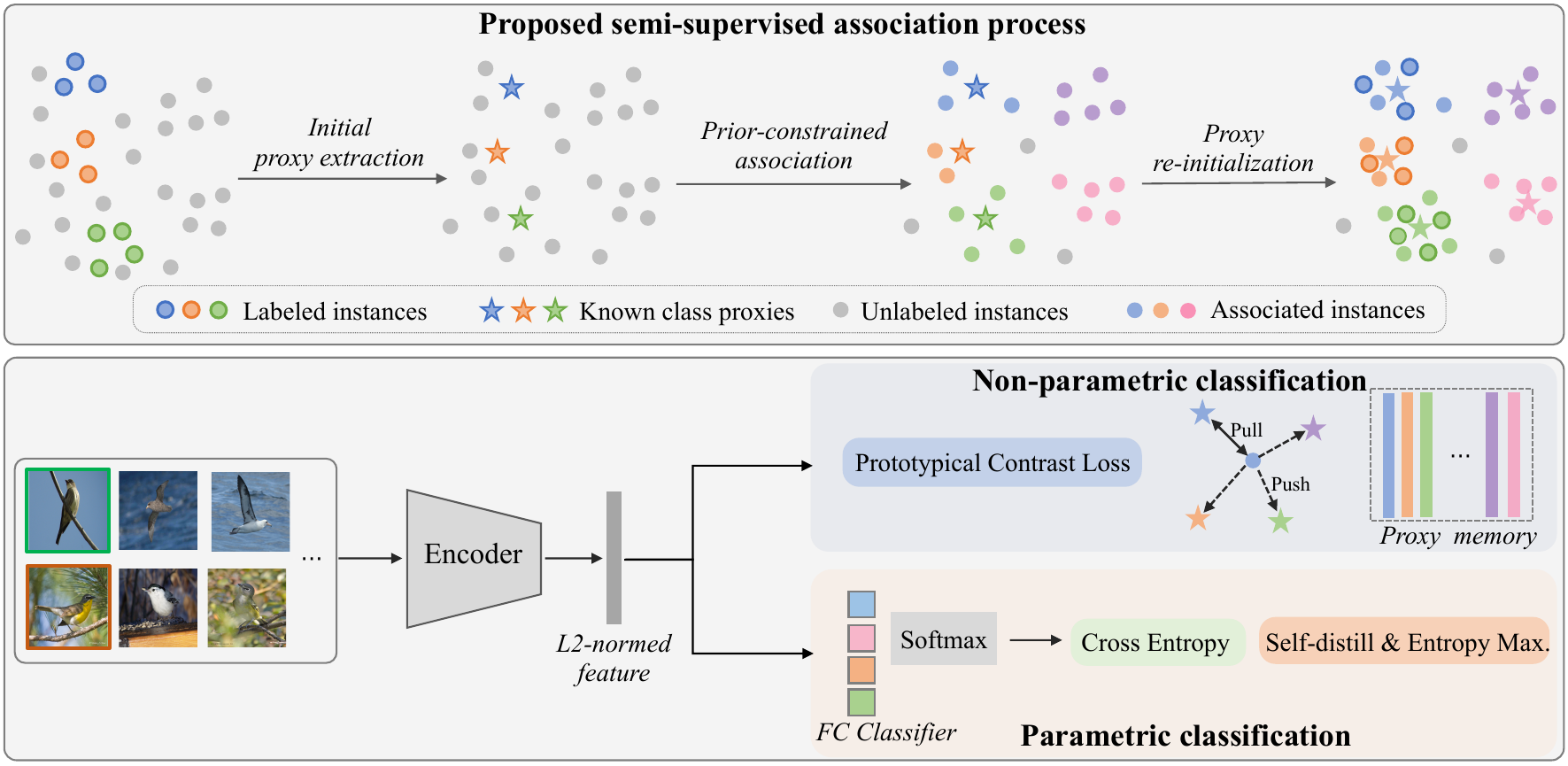} 
\caption{An overview of our method. Non-parametric classification is in the form of prototypical contrastive learning, with prototypes obtained by the proposed semi-supervised association. Parametric classification follows the implementation of SimGCD~\cite{wen2023parametric}.} 
\label{fig_framework}
\end{figure*}

\begin{table}[t]
\centering
\scalebox{0.745}{
\begin{tabular}{ccccc}
\toprule 
Dataset &  CUB & StanfordCars & Aircraft & Herbarium19 \\
\midrule
Estimated class number & 76 & 74 & 27 & 228 \\
G.T. class number & 100 & 98 & 50 & 342 \\
\bottomrule
\end{tabular}
}
\caption{Comparison of class number estimated by pre-clustering refinement + DBSCAN, v.s. ground truth class number in labeled subset of training images.}
\label{intro_table}
\end{table}

However, both pre- and post-clustering refinement neglect the underlying association process. As Figure \ref{fig_intro} shows, inter-class false association can still occur even after simple pre/post-refinement. Simply adopting the standard unsupervised clustering not only fails to address this, but also keeps the incorrect association of unlabeled instances untreated. As an example to verify the problem with pre-clustering refinement, we use the pre-refined inter-instance distance matrix as input to DBSCAN clustering~\cite{ester1996density}, and compare the predicted pseudo class number of labeled subset with the ground truth . In Table \ref{intro_table}, we observe that the clustering predicts much less class number than ground truth, indicating that instances from different labeled classes have been falsely merged. This demonstrates that simply masking the distance of labeled classes during pre-association is insufficient, as these masked instances can still be mis-connected during association.

\subsection{Prior-constrained Greedy Association}
\label{sec:3.3}
In an effort to fix the limitations of the existing association/clustering in GCD, we propose our prior-constrained greedy association algorithm. The initial motivation is that each group should only contain at most one old class. To ensure the constraint is satisfied,  we propose to attend to each step of the association. 
 While associating instance pairs in a distance-ascending order, each step goes through a labeled prior based validity check to make sure one group contains no more than one old class. After such association, the generated instance groups are guaranteed to adhere to the ground truth of labeled subset. An overall pipeline of our association process is shown in the upper part of Figure \ref{fig_framework}.

\begin{itemize}[noitemsep,topsep=0pt] 
 \item 
 \textit{Hybrid feature extraction.} First, we extract feature encodings of all training samples using the backbone network $f$. For the labeled instances, per-category mean feature is calculated as their class representative proxy. Then, our prior-constrained greedy association is performed among the initial known-class proxies and the rest unlabeled instances.

\item
 \textit{Pairwise distance computation.}With the hybrid features of initial proxies and unlabeled instances, we compute their pairwise Jaccard distance~\cite{Zhong2017reranking}, and sort the distances in ascending order. Only pairs whose distance fall below a given threshold $\epsilon$ are kept. Those kept pairs are regarded as the association candidates $P=\{(j_1,j_2),(j_3,j_4),...\}$.

  \item
 \textit{Greedy association with constraint.} In order to incorporate the labeled data prior to constrain the association, we unfold the association in a pair-wise manner. The pseudo algorithm for the association is presented in Alg. 1 of Appendix. Starting from the most similar instance-to-instance (or instance-to-proxy) pair, we obtain the initial grouping $Grp=\{0:(j_1,j_2)\}$. For the next candidate pair $(j_3,j_4)$, we check for conflicts if this candidate pair is to be associated; If the group of instance/proxy $j_3$ contains known category different from the group of instance/proxy $j_4$, this candidate pair will not be associated. The conflict check is performed for every candidate pair in $P$. In this way, every step of association updates the grouping result while still fully respects the ground truth class relations of labeled data. 
 \end{itemize}

\noindent \textbf{\textit{Scaling by subset association.}} The proposed greedy association works by gradually associating reliable instances into cluster groups. This makes it well-suited for fine-grained datasets like Semantic Shift Benchmark~\cite{vaze2021open}, where each category has a moderate number of images.
When scaling to large-scale datasets or scenarios with many images per category, we suggest to perform the association on a randomly sampled subset of unlabeled images. Specifically, at each time of association, a fixed ratio of unlabeled instances are randomly sampled from all unlabeled instances, and then associated along with labeled known proxies. Such subset association enjoys \textit{two benefits}: First, it reduces the number of per-category images for more effective association. Second, for large-scale datasets, the computational cost of association (including pairwise distance computation and greedy association) is significantly reduced.

\subsection{Non-parametric Classification}
\label{sec:3.4}
With the semantic groups predicted by the proposed association, we construct a proxy memory $\mathcal{K} \in R^{C \cdot d}$ representing the feature centroid of each semantic group. At the same time, instance-to-group relation is also estimated from the association. To learn the potential semantic structure, we adopt prototypical contrastive learning paradigm~\cite{wu2018memory} taking the proxy memory as the prototypes. Given an image $x_i$, its $\textit{d}$-dimensional feature $f(x_i)$ is extracted through the backbone network $f$. After $\textit{l}_2$-normalization, the image feature is contrasted with the proxy feature memory, and the prototypical contrastive loss is computed as:
\begin{equation}
\small
\mathcal{L}_{npa} = - \frac{1}{B} \sum_{i=1}^{B} \log \frac{exp(\mathcal{K}[y_i]^T f(x_i)/\tau)}{\sum_{j=0}^{C-1} exp(\mathcal{K}[j]^T f(x_i)/\tau)},
\label{eq_proxy1}
\end{equation}
where $\mathcal{K}[j]$ is the $j$-th entry of the memory, $\tau$ is a temperature factor, $B$ is batch size, and $C$ is the number of proxies in $\mathcal{K}$. The contrastive loss defined in Eq. (\ref{eq_proxy1}) pulls an instance close to the centroid of its group while pushes it away from the centroids of all other groups. This can be seen as non-parametric classification with the external proxy memory serving as the classifier. Feature representation gets enhanced through optimizing the similarity relation between images and the representative proxies.

After each batch forward, the proxy memory features are updated in a moving-average manner~\cite{xiao2017memory} using the online batch image features:
\begin{equation}
\small
\mathcal{K}[\tilde{y}_i] \leftarrow \mu \mathcal{K}[\tilde{y}_i] + (1 - \mu) f_\theta(x_i),
\label{eq:mu}
\end{equation}
where $\mu \in [0,1]$ is an updating rate. To promote the mutual-beneficial effect of representation learning and data association, the two processes are iteratively performed during training. Better representation improves the quality of data association, and the improved association in turn facilitates stronger representation learning.

As shown in our experiments, prototypical contrastive learning driven by the proposed data association can serve as an effective \textit{stand-alone} GCD framework. One advantage of it is that it does not require the prior knowledge of the ground truth class number, offering more flexibility to application. Nevertheless, when the ground truth class number is known, parametric classification~\cite{wen2023parametric} can be integrated as a useful complement to the non-parametric contrastive framework. Next, we briefly introduce the parametric classification, and how the two learning mechanisms can be integrated into one unified framework that produces more powerful model.

\subsection{Joint Non-param. and Param. Classification}
\label{sec:3.6}
\textbf{Parametric Classification.} Following the good practice of semi-supervised learning, a representation parametric classification method SimGCD~\cite{wen2023parametric} performs self-distillation by mining the prediction consistency between augmented views, using the sharpened prediction of one augmented view as the supervision for another view. The total loss $\mathcal{L}_{simgcd}$ includes the cross entropy loss on the labeled data, self-distillation and entropy maximization on both labeled and unlabeled data, as well as batch supervised/self-supervised contrast~\cite{vaze2022generalized}. 

Our association based non-parametric classifier directly learns the globally estimated semantic groups, and presents itself as a complement to parametric classifier. However, combining them in one framework is not so straightforward, as the two types of classifiers are learned with different sampling strategy and at different learning paces. Parametric classifier typically learns at a slower pace, while non-parametric classifier learns faster due to the characteristic of non-parametric classification~\cite{xiao2017memory}.

To improve the effectiveness of joint learning the two classifiers, we propose a two-stage training strategy. In the first ``warm-up'' stage, we only train the backbone network with non-parametric classification, to better prepare the network for joint training. In the second stage, the parametric classifier is added, and a weight $\beta=0.1$ is assigned to the non-parametric classifier loss to balance the learning. The training objective of the second stage is the weighted sum of both classifier loss, \textit{i.e.} $\mathcal{L}= \mathcal{L}_{simgcd} + \beta \mathcal{L}_{npa}$.

\noindent\textbf{Discussion.} 
Association based learning has been previously explored for GCD. For example, DCCL~\cite{pu2023dynamic}, GPC~\cite{zhao2023learning} and PromptCAL~\cite{zhang2023promptcal} all use inter-instance similarity for grouping or pairwise labeling. And prototypical contrastive learning is also adopted by DCCL, GPC and OpenCon~\cite{sun2022opencon} to enhance representation learning. Our method differs from them in two ways: First, we leverage the labeled data prior in GCD task, and let the prior constrain the data association process, ensuring that labeled instances are grouped respecting their label prior. This enables our association to generate much more reliable instance grouping result. Second, we show that non-parametric classification can be effectively combined with parametric classifier to further advance the performance in discovering novel categories.

\section{Experiments}

\subsection{Experimental Setup}
\textbf{Datasets.} We perform experiments on four fine-grained datasets and two generic datasets. Fine-grained datasets include the Semantic Shift Benchmark~\cite{vaze2021open} (CUB-200~\cite{wah2011caltech}, StanfordCars~\cite{krause20133d}, Aircraft~\cite{maji2013fine}), and one long-tailed dataset Herbarium19~\cite{tan2019herbarium}. General datasets include Cifar100~\cite{krizhevsky2009learning} and ImageNet-100~\cite{deng2009imagenet}. Compared to generic recognition, fine-grained datasets are more challenging due to small inter-class variation, and reflect many real-world cases in visual recognition system. Following common settings~\cite{vaze2022generalized}, a subset of all train classes is sampled as the old classes, the rest are new classes. $50\%$ of the images from known classes are used to construct the labeled subset $\mathcal{D_L}$, and the rest images constitute $\mathcal{D_U}$. 

\noindent\textbf{Evaluation metric.} In accordance with standard practice~\cite{vaze2022generalized}, clustering accuracy (ACC) is utilized to evaluate the model performance. During evaluation, the predicted label $\hat{y}$ is compared with the ground truth label $y^\ast$, and ACC is calculated as $ACC=\frac{1}{M}\textstyle \sum_{i=1}^{M} \mathds{1}(y_i^\ast =p(\hat{y}_i)) $, where $M=|\mathcal{D_U}|$, and $p$ is the best permutation of $\hat{y}$ to match the ground truth $y^\ast$.

\begin{table*}[ht]
\centering
\scalebox{0.77}{
\begin{tabular}{l ccc ccc ccc ccc ccc ccc}
\toprule
\multirow{2}{*}{Methods} &  \multicolumn{3}{c}{CUB} & \multicolumn{3}{c}{Stanford Cars}  & \multicolumn{3}{c}{Aircraft}  & \multicolumn{3}{c}{Herbarium19}  & \multicolumn{3}{c}{Cifar100}  & \multicolumn{3}{c}{ImageNet-100} \\
\cmidrule(lr){2-4} \cmidrule(lr){5-7} \cmidrule(lr){8-10} \cmidrule(lr){11-13} \cmidrule(lr){14-16} \cmidrule(lr){17-19} 
& All & Old & New     &  All & Old & New     &  All & Old & New     &  All & Old & New    &  All & Old & New    &  All & Old & New  \\
\midrule
\multicolumn{12}{l}{\textbf{\textit{Ground truth number of classes known}}} \\ 
\midrule
\textit{k}-means~\shortcite{macqueen1967some} & 34.3 & 38.9 & 32.1    & 12.8 & 10.6 & 13.8   & 16.0  & 14.4 & 16.8   & 13.0 & 12.2 & 13.4   & 52.0 & 52.2 & 50.8  & 72.7 & 75.5 & 71.3 \\
RankStats+~\shortcite{han2021autonovel}       & 33.3 & 51.6 & 24.2    & 28.3 & 61.8 & 12.1   & 26.9 & 36.4 & 22.2     & 27.9 & 55.8 & 12.8  & 58.2 & 77.6 & 19.3  & 37.1 & 61.6 & 24.8 \\
UNO+~\cite{fini2021unified}                  & 35.1 & 49.0 & 28.1     & 35.5 & 70.5 & 18.6    & 40.3 & 56.4 & 32.2   & 28.3 & 53.7 & 14.7   & 69.5 & 80.6 & 47.2  & 70.3 & 95.0 & 57.9 \\
GPC~\shortcite{zhao2023learning}                & 52.0 & 55.5 & 47.5     & 38.2 & 58.9 & 27.4    & 43.3 & 40.7 &  44.8   & - & - & -   & 77.9 & 85.0 & 63.0  & 76.9 & 94.3 & 71.0 \\
GCD~\cite{vaze2022generalized}        & 51.3  & 56.6 & 48.7    & 39.0 & 57.6 & 29.9   & 45.0 & 41.1  & 46.9     & 35.4  & 51.0 & 27.0   & 73.0 & 76.2 & 66.5  & 74.1 & 89.8 & 66.3 \\
XCon~\cite{fei2022xcon}                   & 52.1 & 54.3 & 51.0    & 40.5 & 58.8 & 31.7     & 47.7 & 44.4 & 49.4   & - & - & -    & 74.2 & 81.2 & 60.3   & 77.6 & 93.5 & 69.7 \\
DCCL~\shortcite{pu2023dynamic}                 & 63.5 & 60.8 & \underline{64.9}     & 43.1 & 55.7 & 36.2      &- & - & -              &- & - & -   & 75.3 & 76.8 & 70.2   & 80.5 & 90.5 & 76.2 \\
PromptCAL~\shortcite{zhang2023promptcal}    & 62.9 & 64.4 & 62.1   & 50.2 & 70.1 & 40.6     & 52.2 & 52.2 & 52.3   & 37.0 & 52.0 & 28.9  & 81.2 & 84.2 & 75.3   & 83.1 & 92.7 & 78.3 \\
PIM~\shortcite{chiaroni2023parametric}         & 62.7 & 75.7 & 56.2    & 43.1 & 66.9 & 31.6     &- & - & -           & 42.3 & 56.1 & 34.8    & 78.3 & 84.2 & 66.5      & 83.1 & \underline{95.3} & 77.0 \\
$\mu$GCD~\shortcite{vaze2023norep}           & 65.7 & 68.0 & 64.6    & 56.5 & 68.1 & 50.9     & 53.8 & 55.4 & 53.0       & 45.8 & 61.9 & 37.2  &- & - & -     &- & - & -     \\
CMS~\shortcite{Choi2024cms}                       & 68.2 & \textbf{76.5} & 64.0    & 56.9 & 76.1 & 47.6    & 56.0  & 63.4  & 52.3     & 36.4 & 54.9 & 26.4   & \textbf{82.3} & \textbf{85.7} & 75.5   & 84.7 & \textbf{95.6} & 79.2 \\
SimGCD~\shortcite{wen2023parametric}                          & 60.3 & 65.6 & 57.7     & 53.8 & 71.9 & 45.0    & 54.2 & 59.1 & 51.8      & 44.0 & \underline{58.0} & 36.4   & 80.1 & 81.2 & 77.8   & 83.0 & 93.1 & 77.9 \\
SimGCD$ ^\dagger$~\shortcite{wen2023parametric}       & 60.8 & 65.2 & 58.5     & 53.8 & 70.8 & 45.6    & 52.3 & 59.8 & 48.6      & 44.5 & 57.9 & 37.3   & 79.4 & 82.2 & 73.9      & \underline{85.0} & 94.2 & \underline{80.3} \\
\rowcolor{mygray2}  
\textit{Ours} (\textit{param. eval})    & \underline{67.6} & \underline{75.5} & 63.7     & \underline{66.7} & \underline{79.1} & \underline{60.7}            & \textbf{59.5} &  \textbf{67.2} &  \underline{55.6}        & \textbf{47.6} &  \textbf{58.6} &  \textbf{41.7}       & \underline{82.0} & 82.4 & \textbf{81.2}       & \textbf{86.3} & 93.0 & \textbf{83.0}   \\ 
\rowcolor{mygray}  
\textit{Ours} (\textit{nonparam. eval})    & \textbf{72.6} & 75.2 & \textbf{71.2}      & \textbf{72.2} & \textbf{83.4} & \textbf{66.8}      & \underline{58.8} & \underline{64.5} & \textbf{56.0}        &  \underline{46.8} & 57.0 & \underline{41.4}       & 78.5 & 78.9 & \underline{77.8}   & 83.0 & 93.1 & 78.0  \\                  
\midrule
\multicolumn{12}{l}{\textbf{\textit{Ground truth number of classes unknown}}} \\ 
\midrule
GCD~\cite{vaze2022generalized}          & 51.1 & 56.4  & 48.4     & 39.1 & 58.6 & 29.7     & - & - & -              & 37.2 & 51.7 & 29.4      & 70.8 & 77.6 & 57.0   & 77.9 & 91.1 & 71.3 \\
GPC~\shortcite{zhao2023learning}               & 52.0 & 55.5 & 47.5     & 38.2 & 58.9 & 27.4     & 43.3 & 40.7 & 44.8      & 36.5 & 51.7 & 27.9    & 75.4 & \textbf{84.6} & 60.1  & 75.3 & 93.4 & 66.7 \\
PIM~\shortcite{chiaroni2023parametric}     & 62.0 & 75.7 & 55.1      & 42.4 & 65.3 & 31.3     &  - & - & -                      & \underline{42.0} & \underline{55.5} & \underline{34.7}   & 75.6 & 81.6 & 63.6   & \textbf{83.0} & \underline{95.3} & \textbf{76.9} \\
CMS~\shortcite{Choi2024cms}                 & \underline{64.4} & \textbf{68.2}  & \underline{62.4}    & \underline{51.7} & \underline{68.9} & \underline{43.4}      & \underline{55.2}  & \underline{60.6} & \underline{52.4}        & 37.4 & \textbf{56.5} & 27.1  & \textbf{79.6} & \underline{83.2}  & \underline{72.3}        & 81.3 & \textbf{95.6} & 74.2 \\
\rowcolor{mygray}  
\textit{Ours} $^*$                               & \textbf{69.9} & \underline{68.0} & \textbf{70.9}      & \textbf{70.5} & \textbf{77.7} & \textbf{67.0}      & \textbf{56.2} & \textbf{55.3} & \textbf{56.7}      & \textbf{44.8} & 48.5 & \textbf{42.8}          & \underline{77.3} & 78.9 & \textbf{74.1}       & \underline{81.7} & 94.6 & \underline{75.2} \\       
\bottomrule 
\end{tabular}
}
\caption{Performance comparison with SoTA methods. \textit{Ours} (\textit{param. eval}) and \textit{Ours} (\textit{nonparam. eval}): Our full model evaluated with pseudo label predicted by parametric classification logits, or by the proposed association. \textbf{$Ours^*$}: Our model trained with only non-parametric loss. $^\dagger$ denotes reproduced results. Best and second best results are marked by \textbf{Bold} and \underline{underline}.}
\label{compare_sota_table}
\end{table*}

\begin{table*}[ht]
\centering
\scalebox{0.8}{
\begin{tabular}{c ccc | cc | ccc ccc ccc ccc}
\toprule   
& \multicolumn{3}{c|}{ Training}  & \multicolumn{2}{c|}{ Evaluation}  & \multicolumn{3}{c}{CUB} & \multicolumn{3}{c}{Stanford Cars}  & \multicolumn{3}{c}{Aircraft}  & \multicolumn{3}{c}{Herbarium19}  \\
\cmidrule(lr){2-4} \cmidrule(lr){5-6} \cmidrule(lr){7-9} \cmidrule(lr){10-12} \cmidrule(lr){13-15} \cmidrule(lr){16-18} 
& PcA & Param. Cls & 2-stage & A\&A & Param. Cls &  All & Old & New  &  All & Old & New &  All & Old & New &  All & Old & New  \\
\midrule
(1) & &\checkmark & & & \checkmark    & 60.8 & 65.2 & 58.5     & 53.8 & 70.8 & 45.6    & 52.3 & 59.8 & 48.6      & 44.5 & 57.9 & 37.3 \\
(2) & \checkmark & & & \checkmark &   & \underline{69.9} & 68.0 & \underline{70.9}     & \underline{70.5} & 77.7 & \textbf{67.0}     & 56.2 & 55.3 & \textbf{56.7}     & 44.8 & 48.5 & \textbf{42.8} \\ 
\midrule
\rowcolor{mygray2}  
(3) & \checkmark & \checkmark & \checkmark &  & \checkmark  & 67.6 &  \textbf{75.5} & 63.7	& 66.7 & \underline{79.1} &  60.7      & \textbf{59.5} &  \textbf{67.2} &  55.6      & \textbf{47.6} &  \textbf{58.6} &  \underline{41.7} \\
\rowcolor{mygray}  
(4) & \checkmark & \checkmark & \checkmark & \checkmark &   & \textbf{72.6} &  \underline{75.2} & \textbf{71.2}       & \textbf{72.2} &  \textbf{83.4} & \underline{66.8}       & \underline{58.8} & \underline{64.5} & \underline{56.0}    & \underline{46.8} & 57.0 & 41.4  \\
(5) & \checkmark & \checkmark & &  & \checkmark   & 65.1 &  70.7 &  62.3	& 56.8 &  74.3 &  48.4	& 57.1 & 61.0 &  55.2   & 45.1 & \underline{57.4} & 38.5 \\
(6) & \checkmark & \checkmark & & \checkmark &   & 66.9 &  72.0 &  64.4     & 60.0 &  76.6  & 51.9    & 54.8 &  62.8 &  50.9   & 44.1 &  54.3 &  38.6 \\
\bottomrule
\end{tabular}
}
\caption{Ablation study on the main components of our method.  `PcA' denotes the proposed Prior-constrained Association. `Param. Cls' denotes the parametric classifier.  `A\&A' denotes evaluating the model by our Association and Assign.}
\label{ablation_table}
\end{table*}

\noindent\textbf{Implementation Details.} We adopt ViT-B/16~\cite{dosovitskiy2020image} pre-trained on DINO~\cite{caron2021dino} as the backbone network. Following GCD~\cite{vaze2022generalized}, only the last block of the backbone is fine-tuned. Batch size is 128. Learning rate is 0.01 for the first training stage decayed with a cosine annealed schedule, and 0.1 for the second stage of joint training. More implementation details and dataset statistics can be found in Appendix.

\subsection{Comparison with State of The Arts}

In Table \ref{compare_sota_table}, we compare with the state-of-the-art GCD methods under two settings: ground truth number of classes known or unknown. 

\noindent\textbf{Ground truth number of classes known.} Under this setting, we combine the association based non-parametric classification with the parametric classification, where the ground truth class number is utilized as a prior in the latter. In Table \ref{compare_sota_table}, our proposed method achieves state-of-the-art performance on fine-grained datasets, whether using parametric or non-parametric classifier for evaluation. On CUB and Stanford Cars, our method surpasses the previous best method CMS by $4.4\%$ and $15.3\%$ on `All' accuracy. On generic datasets, our method performs on par with SoTA methods. Additionally, we notice a consistent improvement on `New' classes,  proving our method is good at discovering and clustering new categories.

\noindent\textbf{Ground truth number of classes unknown.} Not knowing the ground truth class number is a more practical setting but causes the model learning to be more challenging. In Table \ref{compare_sota_table}, we compare our method with others that does not require the ground truth class number. The results show that with solely non-parametric loss, our method achieves much higher accuracy on all fine-grained datasets, and also delivers consistent performance on generic datasets. The comparisons prove the effectiveness of our method and its flexibility to work under class-unknown setting.

\subsection{Ablation Study}
To investigate how each component affects the model performance, we perform ablation experiments and present the results in Table \ref{ablation_table}.

\noindent\textbf{Effectiveness of the prior-constrained association.} Table \ref{ablation_table} (1) lists the accuracy of training and evaluation with parametric classifier~\cite{wen2023parametric}. Compared with (1), our association based non-parametric classification as denoted by (2) achieves better performance on each dataset. Noticeably on CUB and Stanford Cars, (2) improves the All Acc by $9.1\%$ and $16.7\%$ respectively. 

\noindent\textbf{Effectiveness of joint training.} The full model indicated by (3) and (4) jointly trains with association-based non-parametric classifier and parametric classifier. Compared to (1) and (2), the result in (3) improves the parametric classifier to a large extent, and the accuracy in (4) also consistently boosts over the non-parametric classifier alone. This shows that the joint training is indeed able to benefit both classifiers by mining their complementarity.

\noindent\textbf{Effectiveness of two-stage training.} To validate the necessity of two-stage training, we also provide the results of jointly training non-parametric and parametric classifier in one single stage, as indicated by (5) and (6). Compared with (1) and (2), the single-stage training promotes the accuracy of parametric classifier, but drops the performance of non-parametric classifier,  indicating that a warming-up stage is necessary to better prepare the model for joint training.

\begin{table}[t]
\centering
\scalebox{0.66}{
    \begin{tabular}{c ccc ccc ccc}
    \toprule 
    \multirow{2}{*}{Association}   & \multicolumn{3}{c}{CUB} & \multicolumn{3}{c}{Stanford Cars} & \multicolumn{3}{c}{Aircraft}  \\
    \cmidrule(lr){2-4} \cmidrule(lr){5-7} \cmidrule(lr){8-10}
    &  All & Old & New  &  All & Old & New &  All & Old & New  \\     
    \midrule
    Semi-Kmeans  & 61.0 &  50.6 &  66.2   & 49.7 &  58.9 & 45.2    & 38.2 &  36.2 & 39.2 \\
    Semi-DBSCAN   & 68.3 &  64.8 & 70.1    & 65.2 & 74.7 &  60.6       & 47.4 &  47.8 &  47.2 \\
    \midrule
    \makecell[c]{Semi-DBSCAN \\ w/ constraint}           & \textbf{71.1} & \textbf{72.8} & 70.3      & \underline{68.6} & \underline{77.1} & \underline{64.4}     & \underline{53.7} & \underline{53.6} & \underline{53.8} \\
    \midrule
    Ours w/o constraint          & 67.9 &  61.5 &  \textbf{71.1}    & 65.9 &  72.3 &  62.9      & 47.5 &  52.9 &  44.7 \\
    \midrule
    \rowcolor{mygray}  
    Our association    & \underline{69.9} &  \underline{68.0}  & \underline{70.9}	        & \textbf{70.5} & \textbf{77.7}  & \textbf{67.0}   & \textbf{56.2} & \textbf{55.3} & \textbf{56.7}     \\
    \bottomrule
    \end{tabular}
}
\caption{Comparison of models trained with different association algorithms. Semi-Kmeans is proposed in \cite{vaze2022generalized}. Semi-DBSCAN is based on the clustering algorithm DBSCAN \cite{ester1996density} and inter-class distances among known instances are masked before clustering. Semi-DBSCAN w/ constraint: adds our proposed prior constraint into the semi-DBSCAN clustering. Ours w/o constraint: our association but with the prior constraint removed.}
\label{associ_analyze_table}
\end{table}

\subsection{Analysis on The Proposed Association}
In this subsection, we conduct analysis on the proposed association from different aspects. To focus on the association part, we only adopt the association-based non-parametric classification loss when reporting the performances.

\noindent\textbf{How does the model perform with other association algorithms?}
In Table \ref{associ_analyze_table}, we explore the option of adopting other common clustering algorithms including Semi-Kmeans~\cite{vaze2022generalized} and Semi-DBSCAN~\cite{ester1996density}, both during training and evaluation. From the table, we observe that Semi-DBSCAN shows competitive performance compared to Semi-Kmeans, but stills underperforms the proposed association on all three datasets.

\noindent \textbf{How much contribution does the prior constraint make in association?} 
The prior constraint serves as the key element in our proposed association. We validate its effectiveness in two ways: First, we demonstrate its integration into the classic DBSCAN algorithm which merges the instances greedily. As shown in Table \ref{associ_analyze_table}, adding the prior constraint to Semi-DBSCAN leads to steady improvement on all three datasets, even surpassing our association method on CUB. This highlights the generalizability of the proposed prior constraint. With the constraint incorporated, our association achieves better accuracy than other algorithms on Stanford Cars and Aircraft.

\begin{table}[t]
\centering
\scalebox{0.76}{
    \begin{tabular}{c ccc ccc ccc}
    \toprule 
    \multirow{2}{*}{\makecell[c]{Subset \\ size}}  & \multicolumn{3}{c}{CUB} & \multicolumn{3}{c}{Aircraft} & \multicolumn{3}{c}{Cifar100} \\
    \cmidrule(lr){2-4}  \cmidrule(lr){5-7} \cmidrule(lr){8-10} 
    &  All & Old & New  &  All & Old & New &  All & Old & New  \\     
    \midrule
    100\%      & \textbf{69.9} &  \textbf{68.0}   & \textbf{70.9}   & 50.8 & 55.8 & 48.4                         & 74.4 & \textbf{81.8} & 59.7   \\
    50\%        & 60.8 & 52.5 & 64.9                    & \textbf{56.2} & \textbf{55.3} & \textbf{56.7}            & 77.0 & 79.6 & 71.8 \\
    30\%        & 45.2 & 41.8 & 46.8                      & 48.7 & 38.7 & 53.7                                                    & \textbf{77.3} & 78.9 & \textbf{74.1} \\
    \bottomrule
    \end{tabular}
}
\caption{Comparison of model performance with different subset size for association.}
\label{subset_associate_table}
\end{table}

\noindent\textbf{When is subset association necessary?} For large-scale datasets or when number of images per category is high, we propose to perform the subset association. To verify the effect of subset association, we provide in Table \ref{subset_associate_table} the performance on representative datasets with subset and full-set association. CUB, with an average of 30 images per-category, shows better accuracy with full-set association, which is reasonable as reducing subset size further leads to insufficient data for association. In contrast, Aircraft and Cifar100, with an average of 67 and 500 images per category, benefit from subset association, likely because our association works best with a small amount of representative samples, and involving too many samples per-category brings more noise to association, thus harming the performance.

\begin{figure}[t]
\centering
\includegraphics[width=0.33\textwidth]{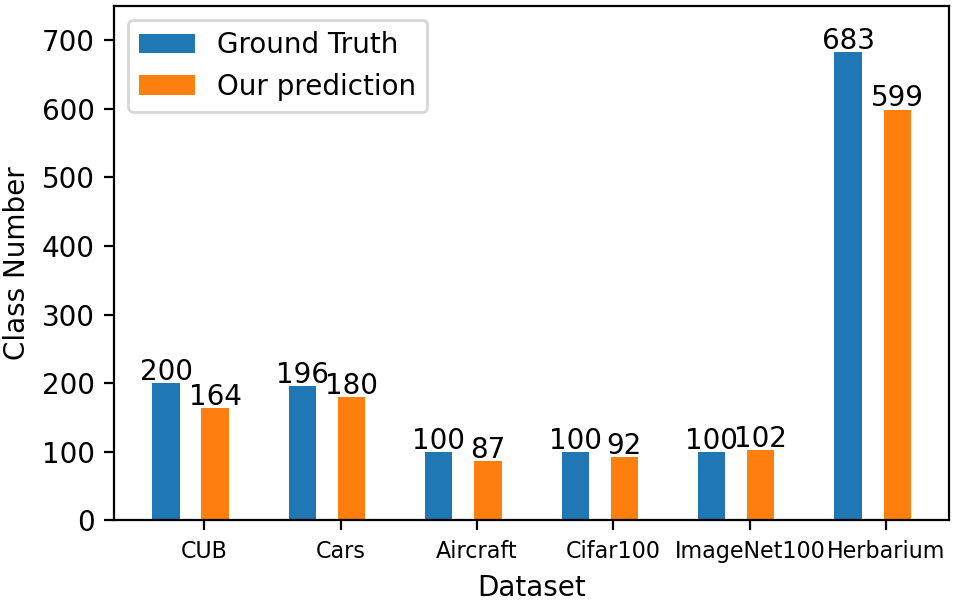}
\caption{Estimated class number on each dataset.}
\label{fig_class_num_pred}
\end{figure}

\noindent\textbf{How well can the proposed method predict the class number?} When the non-parametric classification loss is utilized alone, our method does not require the prior knowledge of ground truth class number. In Figure \ref{fig_class_num_pred}, we compare the estimated \textit{v.s.} ground truth class numbers. In most cases, the association generates fewer pseudo classes than ground truth. Overall, the estimated class number is close to ground truth, with a maximum error rate of $18\%$.

\section{Conclusion}
In this paper, we have proposed a simple yet effective method for generalized category discovery. By mining the labeled data prior under GCD setting, we propose a prior-constrained greedy association algorithm to estimate reliable semantic groups for representation learning. Assisted by the association, the non-parametric prototypical contrastive learning can not only work alone to achieve good performance, but also be effectively integrated with the parametric classifier to mutually benefit each other, leading to further enhanced accuracy. Extensive experiments on multiple benchmarks demonstrate the effectiveness and superiority of the proposed method.

\bibliography{aaai25}

\appendix

\newpage

\section{Appendix}

\subsection{Limitations and impacts}
\subsubsection{Broader Impacts. }
The paper proposes a new method for Generalized Category Discovery (GCD). This task aims at discovering potentially unknown categories in unlabeled images, with the assistance of images from known categories. It has practical applications in recognizing and discovering novel object classes. For example, discovering new animal or plant species with reference to a set of known species. The proposed method improves the clustering performance on multiple standard benchmarks and has the potential to benefit real-world applications.

\subsubsection{Limitations. } 
Our proposed method is targeted at solving GCD task, and fine-grained GCD especially. One limitation of the method is that it is more suitable for datasets with a moderate number of images in each category. For large-scale datasets, the computation of pairwise distance may be slow, and the large number of images per-category may also accumulate association noise which harms the association quality. Although we have proposed subset association as a remedy, this strategy may not fully exploit all the available data. Therefore, one future direction could be to design more scalable association that scales better to all dataset scenarios for GCD.

\subsection{More Implementation Details}

\subsubsection{Model training. } 
ViT-B/16~\cite{dosovitskiy2020image} is adopted as the backbone network. From the network, the 768-\textit{d} class token feature is $\textit{l}_2$-normalized and used as input for parametric classifier. The same feature is also used for non-parametric classification, after forwarded through a batch normalization layer. for For both two training stages, SGD optimizer is utilized, and the default epoch number is 200. On generic datasets, the first stage is trained for 30 epochs only. The hyper-parameters for training the parametric classifier follow SimGCD~\cite{wen2023parametric}. For the non-parametric classifier, the temperature $\tau$ is set as 0.05 and memory update rate $\mu$ is 0.2. The association threshold $\epsilon$ is set as 0.35 for all fine-grained datasets except Herbarium19, on which we use a larger threshold of 0.6 as generic datasets Cifar100 and ImageNet-100. Association is performed in an iterative paradigm at the beginning of every epoch. Due to the number of larger per-category images, subset association is performed on Aircraft, Cifar100 and ImageNet-100 with subset size as $50\%$,  $30\%$ and $30\%$ respectively. On other datasets we adopt whole-set association. For the first training stage when only non-parametric classification is utilized, PK sampler \cite{pu2023dynamic, ge2020self} is adopted for mini-batch sampling. Each batch contains 8 random pseudo classes and 16 instances from each class. For the second stage of joint non-parametric and parametric training, the loss balancing parameter $\lambda$ is set to 0.1, and weighted sampler is adopted following SimGCD~\cite{wen2023parametric}. The experiments are conducted on GTX 1080 and RTX 3090 GPU.

\subsubsection{Model evaluation. } 
The final model after training is used for performance evaluation. When only the association based non-parametric classification is utilized, we use the association result for the pseudo label assignment. Specifically, the association takes the backbone feature of all instances, and generates a number of instance groups. Each instance takes its group index as the pseudo label. For the rest un-associated instances, we assign them to their closest group by comparing feature cosine similarity with all the group center features.

When non-parametric and parametric classifiers are jointly trained, either the association-based assignment or the parametric classifier prediction can be adopted as the pseudo label assignment. For the parametric classifier prediction, we take the \textit{Argmax} index of the logits prediction as the pseudo label, following SimGCD~\cite{wen2023parametric}. 

After obtaining the pseudo label, it is compared to the ground truth label, and clustering accuracy can be computed through Hungarian optimal assignment~\cite{Kuhn1955hungarian}.

\begin{table*}[ht]
\centering
\scalebox{0.9}{
\begin{tabular}{ccccccc}
\toprule 
Dataset & Balanced    &   $\mathcal{Y_L}$  &        $\mathcal{D_L}$     &       $\mathcal{Y_U}$       &    $\mathcal{D_U}$  &    $\#$Average imgs per-category \\ 
\midrule
CUB~\cite{wah2011caltech}               & \cmark     & 100  & 1.5K  & 200  & 4.5K           & 30.0 \\
Stanford Cars~\cite{krause20133d}      & \cmark   & 98   & 2.0K   & 196   & 6.1K          & 41.6 \\
Aircraft~\cite{maji2013fine}                  & \cmark   & 50   & 1.7K  & 100   & 5.0K          & 66.7 \\
Herbarium19~\cite{tan2019herbarium}         & \xmark  & 341  & 8.9K   & 683 & 25.4K  & 50.1 \\
Cifar100~\cite{krizhevsky2009learning}       & \cmark  & 80 & 20K   & 100   & 30K       & 500 \\
ImageNet-100~\cite{deng2009imagenet}      & \cmark  & 50  & 31.9K  & 100 & 95.3K  & 1271 \\
\bottomrule
\end{tabular}
}
\caption{Detailed statistics of each dataset.}
\vspace{-.05in}
\label{dataset_table}
\end{table*}

\subsubsection{Computational analysis of association.} 
During data association, most of the computation overhead are pairwise distance computation and step-wise greedy association.  

\textit{Pairwise distance computation}: Let us assume $C1$ is the number of known categories, $M$ is the number of unlabeled instances, and $d$ is the output feature dimension. The time complexity for pairwise distance computation is $\mathcal{O}((C1+M)(C1+M)d)$. 

\textit{Step-wise greedy association}: After pairwise distance computation, the number of candidate pairs $|P|$ for association depends on the threshold. Normally, a very small proportion of all possible pairs is chosen as candidate pairs, and the computational complexity of the step-wise association process is linear to candidate pair number, \textit{i.e.}, $\mathcal{O}(|P|)$.

The overall computational complexity of association can then be regarded as $\mathcal{O}((C1+M)(C1+M)d) + |P|)$.

\subsubsection{Dataset statistics.}
In Table \ref{dataset_table}, we describe the statistics of the six datasets used in experiments. Of all the six datasets, the first four datasets are fine-grained and the last two are generic datasets. Herbarium19~\cite{tan2019herbarium} is a long-tailed dataset while other datasets are balanced in per-class image distribution. CUB, Stanford Cars and Herbarium19 have an average of less than (or near) 50 images per-category, while Aircraft, Cifar100 and ImageNet-100 have an average of 50 to 1300 images per-category.

\subsection{Pseudo code of the proposed association algorithm}
In Algorithm \ref{our_algorithm}, the pseudo code of the proposed association is provided to facilitate better understanding.

\begin{algorithm}[h]
\scriptsize
\caption{The Prior-Constrained Greedy Association.}
\label{our_algorithm}
\definecolor{codeblue}{rgb}{0.25,0.5,0.5}
\lstset{
  backgroundcolor=\color{white},
  basicstyle=\fontsize{7.2pt}{7.2pt}\ttfamily\selectfont,
  columns=fullflexible,
  breaklines=true,
  captionpos=b,
  commentstyle=\fontsize{7.2pt}{7.2pt}\color{codeblue},
  keywordstyle=\fontsize{7.2pt}{7.2pt},
}
\begin{lstlisting}[language=python]
# Input: distance matrix W, distance threshold thresh, number of known categories C1, number of unlabeled instances M
# Output:  instance-wise pseudo group label grpLabel
grpLabel = -1*ones(C1+M)  # initialize group label 
grpLabel[0: C1] = range(C1)  
count = C1
W[0:C1,0:C1] = thresh+1   # mask dists of old proxies
inds = where(W < thresh)
P = argsort(W[inds])   # sort by distance-ascending 
P = inds[P]
for (i,j) in P:
    # initialize a new group
    if grpLabel[i]==-1 and grpLabel[j]==-1: 
        grpLabel[i], grpLabel[j] = count, count  
        count+=1  
    # associate instances to an existing group  
    elif grpLabel[i]!=-1 and grpLabel[j]==-1: 
        grpLabel[j] = grpLabel[i]              
    elif grpLabel[i]==-1 and grpLabel[j]!=-1:
        grpLabel[i] = grpLabel[j]   
    # merge of two valid groups   
    elif grpLabel[i]!=-1 and grpLabel[j]!=-1 and grpLabel[i]!=grpLabel[j]:
        if grpLabel[i]>=C1 or grpLabel[j]>=C1:  
            minL = min(grpLabel[i], grpLabel[j])
            maxL = max(grpLabel[i], grpLabel[j])
            grpLabel[grpLabel==maxL] = minL
\end{lstlisting}
\end{algorithm}

\begin{figure}[h]
\centering
\includegraphics[width=0.45\textwidth]{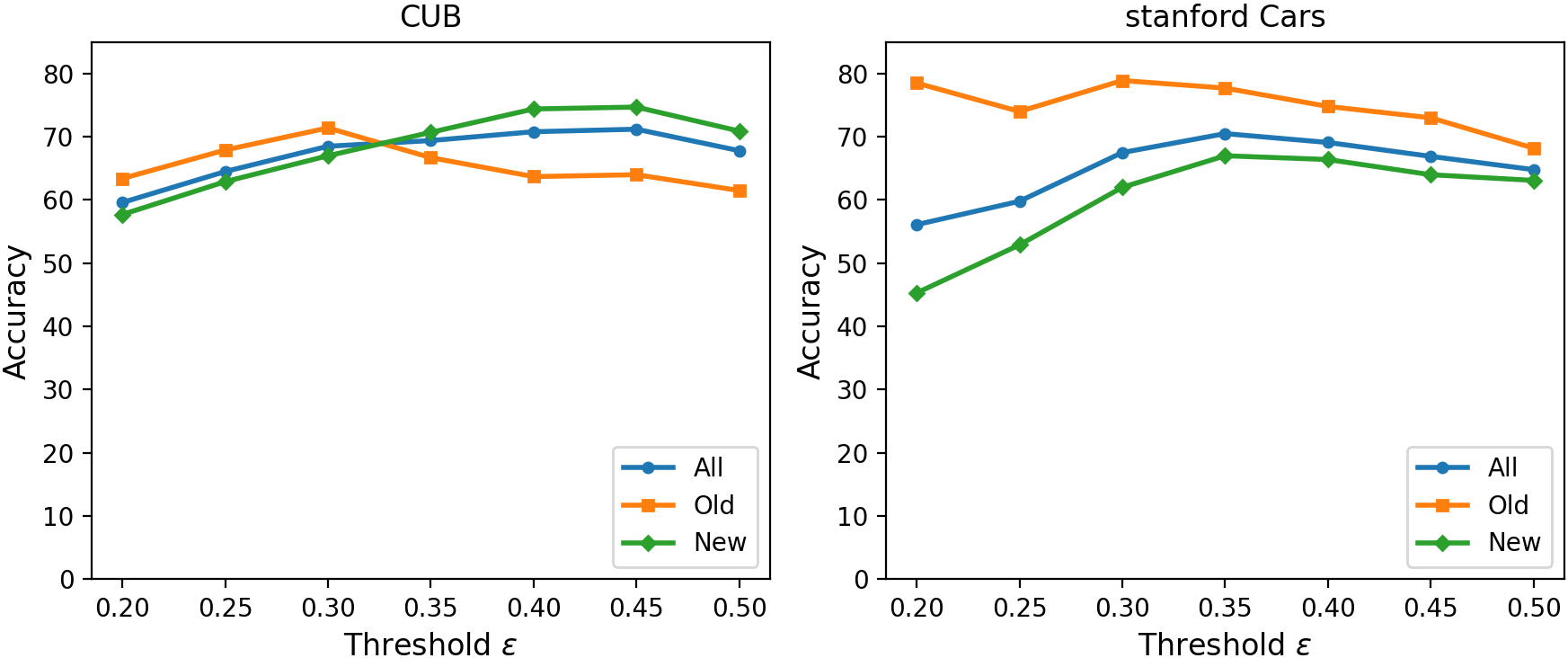} 
\caption{Analysis on association threshold $\epsilon$.}
\label{fig_acc_over_thresh}
\end{figure}

\begin{figure*}[ht]
\centering
\begin{subfigure}{0.3\textwidth}
\centering
\includegraphics[width=1.0\textwidth]{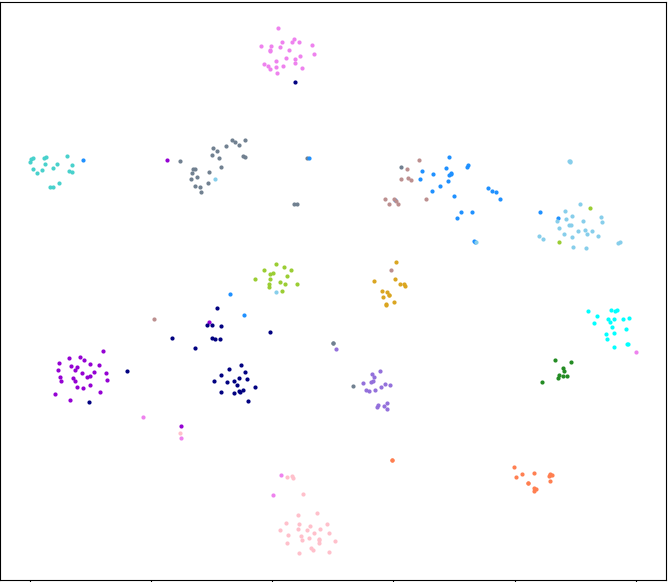} 
\caption{SimGCD model}
\end{subfigure}
\quad
\begin{subfigure}{0.295\textwidth}
\centering
\includegraphics[width=1.0\textwidth]{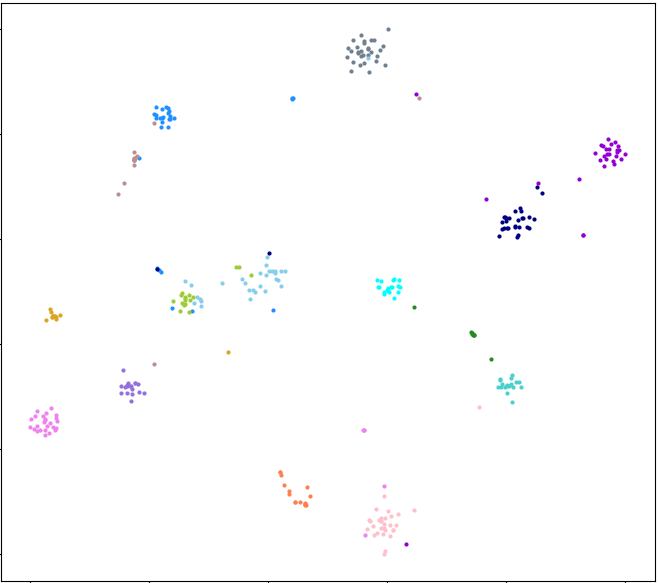} 
\caption{Our model w/ non-param. classifier}
\end{subfigure}
\quad
\begin{subfigure}{0.3\textwidth}
\centering
\includegraphics[width=1.0\textwidth]{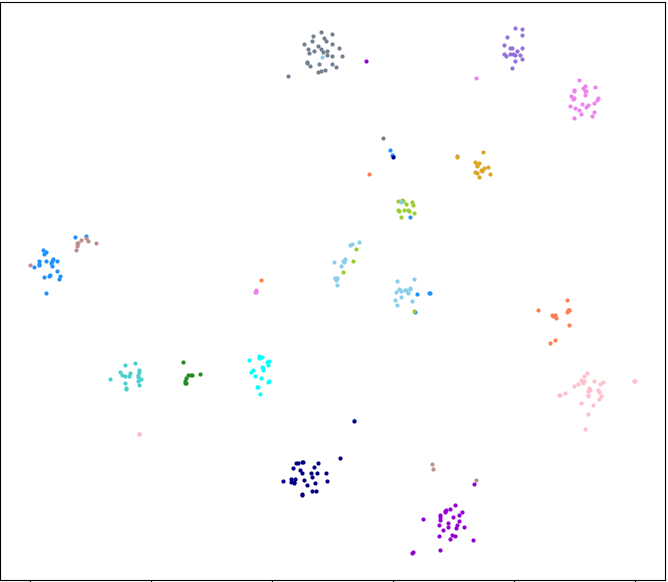} 
\caption{Our full model}
\end{subfigure}
\caption{Visualization of features extracted by different models on CUB dataset.}
\label{fig_visualize}
\end{figure*}

\subsection{More experimental results}

\subsubsection{Analysis on association threshold $\epsilon$.} Figure \ref{fig_acc_over_thresh} plots the model accuracy under varying association thresholds. We observe that accuracy in `New' and `All' classes share a consistent trend, and their peak accuracy appears at a similar threshold. It may be attributed to the `New' classes being unlabeled and harder to cluster, thus influencing the overall accuracy more. Also, the accuracy in `Old' classes favors a smaller threshold compared to `All' and `New' classes. When threshold increases, more instances are assigned to new classes and the bias on `Old' classes gets alleviated. A threshold within the range of [0.3, 0.45] strikes a balance between `Old' and `New' classes.

\subsubsection{Analysis on the loss weight $\beta$. }   
To check the effect of the loss weight $\beta$ during the second stage of joint training, Table \ref{analyze_beta_table} presents the model performance under different $\beta$ values. From the table, we observe that a smaller weight $\beta$ on the non-parametric classifier loss is more beneficial.

\begin{table}[h]
\centering
\scalebox{0.8}{
    \begin{tabular}{c ccc ccc}
    \toprule 
    \multirow{2}{*}{$\beta$}  & \multicolumn{3}{c}{\textit{param. eval}} & \multicolumn{3}{c}{\textit{nonparam. eval}} \\
    \cmidrule(lr){2-4}  \cmidrule(lr){5-7} 
    &  All & Old & New  &  All & Old & New  \\     
    \midrule
    0.05                    & 71.9 & 83.0 & 66.5        & 65.5 & 80.9 & 58.0 \\
    0.1                      & \textbf{72.2} & \textbf{83.4} & \textbf{66.8}        & \textbf{66.7} & 79.1 & \textbf{60.7} \\
    0.2                     & 71.5 & 82.3 & 66.3         & 63.6 & \textbf{81.5} & 54.9 \\ 
    0.5                     &  71.4 & 82.8 & 65.9         & 64.2 & 77.8 & 57.8 \\
    1                       & 70.9 & 82.4 & 65.3           & 65.6 & 80.3  & 58.5 \\
    \bottomrule
    \end{tabular}
}
\caption{Analysis of loss weight $\beta$ on Stanford Cars dataset.}
\label{analyze_beta_table}
\end{table}

\subsubsection{Accuracy evaluated by association before and after training. } 
To gain a more clear idea of the performance boost, Table \ref{before_after_table} compares the model accuracy before and after training, evaluated by association-and-assign. Before training, the DINO-pretrained backbone is utilized to extract image features for association, reflecting the model's initial ability to recognize and cluster images. As shown in Table \ref{before_after_table}, the initial accuracy is generally lower for fine-grained datasets, and higher for generic datasets, indicating that the DINO-pretrained feature is better at generic recognition than fine-grained recognition. After training, the accuracy is significantly improved on fine-grained dataset, especially Stanford Cars where the `All' Acc is lifted from 12.9 to 70.5. The before$\&$after comparison demonstrates the effectiveness of association-based training to improve representation.

\begin{table}[h]
\centering
\scalebox{0.9}{
    \begin{tabular}{c ccc ccc}
    \toprule 
    \multirow{2}{*}{Dataset}  & \multicolumn{3}{c}{Before training} & \multicolumn{3}{c}{After training} \\
    \cmidrule(lr){2-4}  \cmidrule(lr){5-7} 
    &  All & Old & New  &  All & Old & New  \\     
    \midrule
    CUB                    & 35.3 & 49.4 & 28.2        & 69.9 & 68.0 & 70.9 \\
    Stanford Cars      & 12.9 & 19.8 & 9.6           & 70.5 & 77.7  & 67.0 \\
    Aircraft                & 15.0 & 13.8 & 15.5         & 56.2 & 55.3 & 56.7 \\
    Herbarium19       & 14.4 & 18.0 & 12.4         & 44.8 & 48.5 & 42.8  \\
    Cifar100              & 53.5 & 60.0 & 40.6          & 77.3 & 78.9 & 74.1 \\
    ImageNet-100      & 79.2 & 89.1 & 74.3          & 81.7 & 94.6 & 75.2 \\
    \bottomrule
    \end{tabular}
}
\caption{Accuracy evaluated by association before and after training.}
\label{before_after_table}
\end{table}

\subsubsection{Error bars for our main results with unknown GT class number.}
In Table \ref{error_bar_table}, we present the error bar result of our method under unknown GT class number (i.e. models trained with only non-parametric classifier loss). Each result is obtained from three independent runs.

\begin{table}[h]
\centering
\scalebox{0.9}{
    \begin{tabular}{c ccc}
    \toprule 
     Dataset &  All & Old & New  \\     
    \midrule
    CUB                          & 69.9±0.6 & 68.0±1.2 & 70.9±0.3 \\
    Stanford Cars             & 70.5±0.4 & 77.7±0.8  & 67.0±0.3 \\     
    Aircraft                      & 56.2±0.6 & 55.3±1.2 & 56.7±0.3 \\        
    Herbarium19             & 44.8±0.3 & 48.5±0.7 & 42.8±0.2  \\
    Cifar100                     & 77.3±0.7 & 78.9±0.1 & 74.1±2.1 \\
    ImageNet-100             & 81.7±0.5 & 94.6±0.1 & 75.2±0.7 \\
    \bottomrule
    \end{tabular}
}
\caption{Error bars of our method on each dataset.}
\label{error_bar_table}
\end{table}

\subsubsection{Coping with lower ratio of labeled subset. } 
The default dataset setting of GCD is to set known class ratio as $50\%$, and randomly select $50\%$ images from the known classes as labeled. To testify the robustness of the proposed method, we create more challenging settings with varying ratio $\{0.1, 0.25, 0.5\}$ of known category or images per known category. The experimental results are presented in Table \ref{low_ratio_table}. From the table, we observe that: 1) Our method is able to maintain a relatively good performance when reducing the known class ratio or samples per known class ratio to 0.25. Compared to SimGCD, our method experiences much less accuracy loss when coping with less known classes or known samples per-class. 2) our method is more robust to the decrease in samples-per-known-class, compared to the decrease in number of known classes. This indicates the applicability of our method to data scenarios where rare-category images are hard to collect and labeled in advance.

\begin{table}[h]
\centering
\scalebox{0.7}{
    \begin{tabular}{ccc ccc ccc}
    \toprule 
    \multirow{2}{*}{Method} & \multirow{2}{*}{Class ratio}   &  \multirow{2}{*}{Sample ratio}   & \multicolumn{3}{c}{CUB} & \multicolumn{3}{c}{Stanford Cars} \\
    \cmidrule(lr){4-6}  \cmidrule(lr){7-9} 
    & & &  All & Old & New  &  All & Old & New  \\     
    \midrule
    Ours & 0.1  & 0.5            &  64.4 &  59.7   & 64.6       &  50.4 &  59.8   & 49.9 \\
    Ours & 0.25  & 0.5          &  67.5 &  56.1   & 69.4              &  59.3 &  76.2   & 56.5 \\
    \midrule
    Ours & 0.5  & 0.1             &  65.3 &  67.6   & 63.2                     &  59.7 &  62.6   & 57.2 \\
    Ours & 0.5  & 0.25           &  69.1 &  \textbf{71.5}   & 67.3       &  \underline{66.7} &  72.8   & \underline{62.3} \\
    \midrule
    Ours & 0.25  & 0.25          &  65.2 &  68.6   & 64.3                     &  59.5 &  \textbf{79.3}   & 54.6 \\
    SimGCD & 0.25  & 0.25   &  39.3 &  30.5   & 41.5                       &  14.9 &  31.3   & 10.8 \\
    Ours & 0.5  & 0.5             &  \textbf{69.9} &  \underline{68.0}   & \textbf{70.9}             &  \textbf{70.5} &  \underline{77.7}   & \textbf{67.0} \\
    SimGCD & 0.5  & 0.5      &  60.3 &  65.6   & 57.7            &  53.8 &  71.9   & 45.0 \\
    \bottomrule
    \end{tabular}
}
\caption{Performances under more challenging data split settings. `Class ratio' is short for Known Class Ratio, `Sample ratio' is short for Samples Per Known Class Ratio.}
\label{low_ratio_table}
\end{table}

\subsubsection{Feature visualization. }
In Figure \ref{fig_visualize}, we visualize the image features extracted by (a) SimGCD model, (b) our model with only non-parametric classifier, and (c) our full model, respectively. The images are from 15 randomly chosen categories in CUB dataset. First by looking at (a) and (b), it is clear that compared to SimGCD,  our association-based non-parametric classifier produces more compact features within category, and inter-category features are also more separable. Comparing (b) and (c), we see that the intra-category compactness is retained, and some confusing categories are better separated.

\end{document}